\documentclass[letterpaper, 10pt, conference]{ieeeconf}
\IEEEoverridecommandlockouts                              
\usepackage{cite}        
\usepackage{amsmath,amssymb,amsfonts} 
\usepackage{algorithm}   
\usepackage{algpseudocode}
\usepackage{graphicx}   
\usepackage{textcomp}
\usepackage{xcolor}
\usepackage{booktabs}    
\usepackage{tabularx}
\usepackage{multirow}
\usepackage{float}
\usepackage{makecell} 
\usepackage{siunitx}  

\usepackage[pagebackref=false,breaklinks=true,colorlinks,bookmarks=false,allcolors=black]{hyperref}

\usepackage{bm}

\title{\LARGE \bf
Learning to Design City-scale Transit Routes
}

\author{Bibek Poudel$^{1}$ and Weizi Li$^{1}$%
\thanks{$^{1}$The authors are with the University of Tennessee, Knoxville, USA.
        {\tt\small bpoudel3@vols.utk.edu, weizili@utk.edu}}%
}

\begin{document}

\maketitle
\thispagestyle{empty} 
\pagestyle{empty}     


\begin{abstract}
Designing efficient transit route networks is an NP-hard problem with exponentially large solution spaces that traditionally relies on manual planning processes. We present an end-to-end reinforcement learning (RL) framework based on graph attention networks for sequential transit network construction. To address the long-horizon credit assignment challenge, we introduce a two-level reward structure combining incremental topological feedback with simulation-based terminal rewards. We evaluate our approach on a new real-world dataset from Bloomington, Indiana with topologically accurate road networks, census-derived demand, and existing transit routes. Our learned policies substantially outperform existing designs and traditional heuristics across two initialization schemes and two modal-split scenarios. Under high transit adoption with transit center initialization, our approach achieves 25.6\% higher service rates, 30.9\% shorter wait times, and 21.0\% better bus utilization compared to the real-world network. Under mixed-mode conditions with random initialization, it delivers 68.8\% higher route efficiency than demand coverage heuristics and 5.9\% lower travel times than shortest path construction. These results demonstrate that end-to-end RL can design transit networks that substantially outperform both human-designed systems and hand-crafted heuristics on realistic city-scale benchmarks.
\end{abstract}

\section{Introduction}
\label{sec:intro}

Urban population is projected to account for $60\%$ of the world's population by $2030$~\cite{unwup2018}. This rapid urbanization places unprecedented pressure on transit infrastructure, which forms the backbone of urban mobility and enables billions of daily trips worldwide. In the United States alone, transit systems recorded $7.7$ billion trips in $2024$, generating $94$ billion in economic activity~\cite{apta2025, apta2024factbook}. Designing transit networks, which involves selecting route paths, setting their service frequencies, and determining how routes interconnect, is formalized as the \emph{Transit Route Network Design Problem} (TRNDP). The problem is NP-hard, primarily due to combinatorial explosion, i.e.,  as the network size increases, the number of possible route configurations grows exponentially. For instance, the Bloomington network studied in this work yields approximately $9.5\times 10^{115}$ possible route combinations to design $16$ routes of length $14$ each (see supplementary material section~\ref{app:searchspace} for calculation). Hence, exhaustive optimization or systematic evaluation of this vast solution space is computationally intractable, even for modestly sized cities~\cite{newell1979some}. As a result, most transit networks today are designed through manual planning processes and often remain far from optimal~\cite{fan2004optimal, schmidt2024planning}.

TRNDP exhibits additional complexity that differentiates it from classical routing problems such as the Traveling Salesman Problem (TSP) or the Vehicle Routing Problem (VRP). TSP optimizes a single tour visiting each node exactly once, while VRP assigns vehicles to serve fixed customers within defined capacity and time limits. In contrast, TRNDP involves selecting a set of overlapping transit routes embedded within a shared network, determining stop sequences, and setting service frequencies to manage many-to-many passenger flows, including transfers~\cite{ceder2016public}. Designing such routes requires balancing the objectives of passengers, who value higher coverage, shorter waiting times, and direct connections, with those of operators, who must adhere to constraints on fleet size, budget, and service reliability. For example, increasing route frequencies can reduce passenger waiting times and crowding, but it simultaneously raises operating costs and fleet requirements. Further, minor adjustments to one transit route during the design process can propagate across the entire network, rerouting passenger flows, changing transfer volumes, and shifting loads on distant segments~\cite{kepaptsoglou2009transit}. These complex interactions produce a high-dimensional, nonconvex optimization landscape, motivating the necessity for data-driven approaches.

\begin{figure*}[t!]
    \centering
    \includegraphics[width=0.99\linewidth]{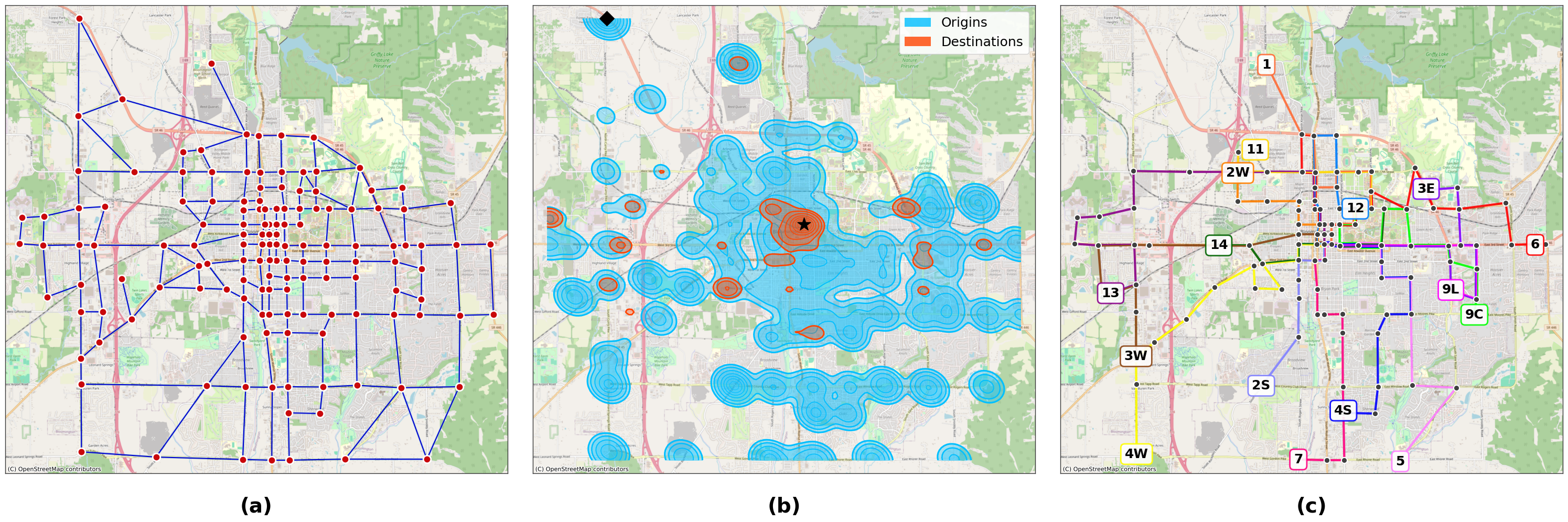}
    \vspace{-7pt}
    \caption{(a) The Bloomington transportation network with $143$ nodes and $243$ edges. (b) Spatial distribution of trip origins (blue) and destinations (red) showing concentrated activity areas with highest demand origin of $344$ veh/hr from node $1$ ($\blacklozenge$), and highest demand destination of $1,681$ veh/hr to node $129$ ($\bigstar$). (c) The $16$ transit routes currently operating in the city.}
    \vspace{-12pt}
    \label{fig:bloomington}
\end{figure*}


Prior work on TRNDP has largely relied on metaheuristic algorithms such as genetic algorithms~\cite{mumford2013new}, simulated annealing~\cite{zhao2006simulated}, and bee colony optimization~\cite{nikolic2013transit}. While these methods have demonstrated success on benchmark networks, they depend heavily on manually designed heuristics and operators for generating and modifying routes, which require significant domain expertise~\cite{kepaptsoglou2009transit}. Recent work has begun to explore data-driven approaches such as reinforcement learning (RL), showing promise in learning adaptive policies for network design~\cite{yoo2023reinforcement}. However, most RL-based methods have been evaluated only on small synthetic networks that fail to capture the rich topological structure and demand patterns of real cities~\cite{heyken2019adaptive, vermeir2021exact}. Additionally, prior RL approaches have struggled with the sparse reward problem inherent in long-horizon sequential network design, where feedback is only available after complete routes are constructed, making credit assignment difficult and learning inefficient~\cite{sutton2018reinforcement}. Even when RL is applied to transit design, recent work has focused on learning heuristics to guide metaheuristic algorithms rather than developing end-to-end policies~\cite{holliday2025learning}.

We address these limitations by developing an end-to-end RL framework that combines graph attention~\cite{brody2021attentive} with proximal policy optimization to learn policies for sequential route network design. Our contributions are:

\begin{itemize}
    \item We introduce a new real-world dataset with topologically accurate road network from Bloomington, Indiana ($143$ nodes, $243$ edges), census block level Origin-Destination (OD) demand, and $16$ existing transit routes.
    \item We extend a mesoscopic traffic simulator with a RL training framework by adding a transit network optimization layer. The framework incorporates a two-level reward function that combines lightweight proxy during route construction with simulation-based terminal rewards to address long-horizon credit assignment.
    \item We propose a scalable policy architecture based on graph attention where the parameters remains independent of input network size and demonstrate its effectiveness with improvements over real-world designs and traditional heuristics across both passenger and operator metrics.
\end{itemize}


Our results show that the trained policy achieves strong performance across multiple metrics on the Bloomington network. Under high transit adoption ($\alpha=1.0$), our approach achieves a service rate of $73.1\%$, representing a $25.6\%$ improvement over the existing real-world network, while reducing average wait time by $30.9\%$ and improving bus utilization by $21.0\%$. Under mixed mode share where a fraction of travelers use public transit ($\alpha=0.3$), the method delivers $68.8\%$ higher route efficiency than a greedy heuristics such as demand coverage, while reducing wait times by $27.3\%$ and achieving $5.9\%$ lower travel times than shortest path heuristic. These results validate the effectiveness of end-to-end RL for real-world transit route network design. We make our dataset and code publicly available.

\section{Related Work}
\label{sec:related_work}
 
TRNDP is commonly framed as a multi-objective, NP-hard combinatorial optimization problem that balances passenger outcomes such as coverage and travel time with operator goals such as limiting fleet size and avoiding unnecessary overlap of services~\cite{kepaptsoglou2009transit, duran2022survey, schmidt2024planning}. Due to the problem's computational complexity, researchers have focused on developing tractable problem instances and solution methods. Early work defined small benchmark instances, most notably the Mandl network (with $15$ nodes), remains a standard testbed for many algorithmic studies~\cite{mandl1980evaluation, mumford2013new}. To search this large design space, a long line of research has developed metaheuristics, including evolutionary approaches~\cite{mumford2013new, fan2010metaheuristic}, simulated annealing~\cite{zhao2006simulated}, bee colony optimization~\cite{nikolic2013transit}, and demand-driven route generation schemes~\cite{kilicc2014demand, yoon2024sequential}. Although these methods can produce good solutions on widely used benchmarks, they rely heavily on hand-crafted move operators, penalty functions, and parameter schedules whose performance must be tuned case-by-case~\cite{huang2019survey, duran2022survey}. In addition, most evaluations rely on synthetic or strongly aggregated networks with simplified demand, which makes it difficult to capture how route layout, congestion, transfers, and frequency interact in a realistic traffic environment.

Recent advances in neural combinatorial optimization~\cite{bello2016neural, vinyals2015pointer} have demonstrated that deep RL can learn effective policies for vehicle routing~\cite{kool2019attention, nazari2018reinforcement} and graph-based assignment~\cite{dai2017learning}, motivating their application to transit network design. For transit network design specifically, RL-based approaches take two forms: hybrid frameworks that use learned policies to guide evolutionary search~\cite{holliday2023augmenting, holliday2025learning, holliday2024neural}, and pure end-to-end methods that directly construct complete networks~\cite{yoo2023reinforcement, darwish2020optimising, li2023transit}. Despite these differences, both approaches share important limitations. First, their validation has been confined almost entirely to small synthetic benchmarks. Second, they typically provide reward only after an entire network has been constructed, even though routes are built node-by-node; this training setup yields sparse and delayed feedback and makes credit assignment across the long decision horizon particularly difficult~\cite{sutton2018reinforcement, yoo2023reinforcement}. Additionally, hybrid methods remain fundamentally constrained by the exploration strategies and convergence properties of their underlying metaheuristic optimizers, which ultimately bound achievable performance. In this work, we develop an end-to-end framework that makes use of a two-level reward function to provide intermediate structural feedback during route construction alongside terminal rewards from full traffic simulation. Further, unlike prior works confined to synthetic benchmarks, we evaluate on a real-world network from Bloomington, Indiana with accurate road topology and census-based demand.

\section{Methodology}
\label{sec:method}


\subsection{Real-world Network and Demand}
We propose a new real-world transit design dataset that includes: (i) a topologically correct road graph with $143$ nodes and $243$ bidirectional edges, extracted from the street network in Bloomington, Indiana; (ii) a block-level OD demand matrix extracted from U.S. Census data; and (iii) the $16$ existing Bloomington Transit routes~\cite{bloomingtontransit_gtfs}. Unlike many benchmarks that rely on synthetic networks and demand, our dataset provides greater realism in the design and evaluation of transit networks. Figure~\ref{fig:bloomington} visualizes the network, the spatial distribution of origins and destinations, and the current routes. For processing details and assumptions, see the supplementary material section~\ref{app:data_processing}.

\begin{algorithm}[t]
\caption{Transit Route Network Design}
\label{alg}
\begin{algorithmic}[1]
    \State \textbf{Input:} Network Graph $G=(V,E)$, OD matrix $D$, Directed edges $\mathcal{I}$, Edge features $Z$, Number of routes $K$, Max route length $L_{\max}$, Update frequency $M$, Policy $\pi$
    \State \textbf{Output:} routes $\Pi=\{r_1,\dots,r_K\}$
    \State \textbf{Initialize:} Buffer $B\gets\varnothing$, Routes $\Pi\gets\varnothing$, Completed nodes $V_{\mathrm{cmp}}\gets\varnothing$
    \For{each episode}
        \For{$k=1$ \textbf{to} $K$}
            \State $r_k\gets[\,]$, $V_{\mathrm{cur}}\gets\varnothing$ 
            \While{$|r_k|<L_{\max}$}
                \State $\mathcal{C}_t\gets$ valid one hop neighbors of the frontier
                \If{$\mathcal{C}_t=\varnothing$}
                    \State \textbf{break} \Comment{Automatic termination}
                  \EndIf
                \State $\mathsf{Form}~X_t$ using $(V_{\mathrm{cur}},\,V_{\mathrm{cmp}})$
                \State $s_t \gets (X_t,\mathcal{I},Z)$
                \State $m_t \gets \mathsf{Mask}(\mathcal{C}_t,n)$ \Comment{Eq. \eqref{eq:masked}}
                \State $\mathsf{Sample}~a_t\sim \pi_\theta(\cdot\mid s_t)$ using $m_t$
                \State $\mathsf{Append}~a_t$ to $r_k$
                \If{$|r_k|~{<}~L_{\max}$}
                    \State $\mathsf{Reward}~r_t\gets \mathcal{R}_{\text{partial}}$ \Comment{Eq.~\eqref{eq:partial}}
                \Else
                    \State $\mathsf{Reward}~r_t\gets \mathcal{R}_{\text{final}}$ \Comment{Eq.~\eqref{eq:final}}
                \EndIf
                \State $\mathsf{Push}~(s_t,a_t,r_t,m_t)$ to $B$
                \State $V_{\mathrm{cur}}\gets$ nodes in $r_k$ 
                \If{$|B|\ge M$} \State $\theta\gets\mathsf{UpdatePolicy}(B)$;\ $B\gets\varnothing$ \EndIf
            \EndWhile
            \State $V_{\mathrm{cmp}}\gets V_{\mathrm{cmp}}\cup V_{\mathrm{cur}}$
        \EndFor
        \State $\Pi\gets \Pi\cup\{r_k\}$  
    \EndFor    
    \State \Return $\Pi$
\end{algorithmic}
\end{algorithm}

\subsection{Markov Decision Process}
We formulate the TRNDP as a finite-horizon Markov decision process defined by the tuple $(\mathcal{S}, \mathcal{A}, P, \mathcal{R}, \gamma)$ on a city graph $G=(V,E)$ with $n=\lvert V\rvert$ nodes and an OD matrix $D\in\mathbb{R}_{\ge 0}^{n\times n}$, where $\mathcal{S}$ is the state space, $\mathcal{A}$ is the action space, $P: \mathcal{S} \times \mathcal{A} \to \mathcal{S}$ is the transition function, $\mathcal{R}: \mathcal{S} \times \mathcal{A} \to \mathbb{R}$ is the reward function, and $\gamma \in [0, 1]$ is the discount factor. In each episode, the agent sequentially designs $K$ routes $\Pi=\{r_1,\ldots,r_K\}$, each with a maximum length of $L_{\max}$ nodes, by extending one route at a time until all routes are finalized. At each step, the agent extends the current route $r_k$ by one node.


We seek to learn a policy that balances passenger and operator objectives. From the passenger perspective, the policy maximizes network coverage and service rate by connecting as many origin-destination pairs as possible while minimizing total travel time. From the operator perspective, it penalizes excessive route overlaps and encourages high service rate. The learning objective is:
\begin{equation}
\label{eq:rl-objective}
\max_{\theta}\quad J(\theta)\;=\;\mathbb{E}_{\pi_\theta}\Bigl[\sum_{t=0}^{T-1}\gamma^t\,\mathcal{R}(s_t,a_t)\Bigr],
\end{equation}
where $\pi_\theta$ is the policy parameterized by $\theta$, $s_t$ and $a_t$ are the state and action at time $t$, and $T$ is the episode horizon ($T = 224$ for our setting with $K=16$ routes of maximum length $L_{\max}=14$). The policy is trained using PPO~\cite{schulman2017proximal} with $\gamma=0.99$ and generalized advantage estimation~\cite{schulman2015high}. Algorithm~\ref{alg} presents the complete design procedure under the following constraints (formal definitions in the supplementary material section~\ref{sec:definitions}):

\begin{enumerate}
\item \emph{Fixed route count}: The designed solution comprises exactly $K$ routes, i.e., $\lvert \Pi \rvert = K$.
\item \emph{Graph connectivity}: Each route $r_k$ forms a walk on $G$, where consecutive nodes must share an edge; formally, for any adjacent pair $(v_i, v_{i+1})$ in $r_k$, we require $\{v_i, v_{i+1}\} \in E$, specified by the user.
\item \emph{Bidirectional operation}: Each route $r_k = (v_1, v_2, \ldots, v_\ell)$ is operated in both directions, with buses traversing the forward sequence $v_1 \to v_2 \to \cdots \to v_\ell$ and the reverse sequence $v_\ell \to v_{\ell-1} \to \cdots \to v_1$, enabling bidirectional O-D connectivity along the route.
\item \emph{Simple paths}: Routes are self-avoiding walks with no node appearing more than once within a single route.
\item \emph{Length bounds}: Each route satisfies $L_{\min} \le \lvert r_k \rvert \le L_{\max}$, where $\lvert r_k \rvert$ is the number of nodes in route $r_k$.
\item \emph{Sequential construction}: Routes are built by extending one node at a time from a frontier node (the most recently added node). At each step, the action space is restricted to the one-hop neighbors of the frontier that have not yet been visited in the current route being constructed. A route terminates upon reaching $L_{\max}$ or when no valid neighbors remain.
\end{enumerate}


\subsubsection{State} 
The state ($s_t$) encodes the static network and evolving design context on $G$ with $s_t \;=\; \bigl(X_t,\ \mathcal{I},\ Z\bigr).$ 

\paragraph{Node features ($X_t\in\mathbb{R}^{n\times 16}$)} Let $V_{\mathrm{cur}}$ be the nodes already placed on the route under construction at time $t$, $V_{\mathrm{cmp}}$ the union of nodes across completed routes, and $V_{\mathrm{core}}=V_{\mathrm{cur}}\cup V_{\mathrm{cmp}}$. Let $\mathcal{C}_t$ be the set of admissible candidates at time $t$, i.e., the one-hop neighbors of the current frontier that are not already in the route $V_{\mathrm{cur}}$. The node features for node $i$ can then be classified into six groups:


\begin{itemize}
    \item Geometry and connectivity: normalized $(x_i,y_i)$ coordinates and node degree.
    \item OD marginals:
    \begin{align}
      d_{\text{out}}(i) &= \sum_{j=1}^{n} D_{ij}, & 
      d_{\text{in}}(i)  &= \sum_{j=1}^{n} D_{ji}. \label{eq:od_marginals}
    \end{align}
    \item Candidate demand between $i$ and the current route (nonzero only if $i\in\mathcal{C}_t$):
    \begin{align}
      a^{\text{cand}}_{i\to \mathrm{cur}} &= \mathbf{1}\{i\in\mathcal{C}_t\}\sum_{j\in V_{\mathrm{cur}}} D_{ij}, \label{eq:cand_out}\\
      a^{\text{cand}}_{i\leftarrow \mathrm{cur}} &= \mathbf{1}\{i\in\mathcal{C}_t\}\sum_{j\in V_{\mathrm{cur}}} D_{ji}. \label{eq:cand_in}
    \end{align}
    \item Designed network demand at $i$ with respect to nodes already in any designed route (nonzero only if $i\in V_{\mathrm{core}}$):
    \begin{align}
      a^{\text{core}}_{i\to \mathrm{core}} &= \mathbf{1}\{i\in V_{\mathrm{core}}\}\sum_{j\in V_{\mathrm{core}}} D_{ij}, \label{eq:core_out}\\
      a^{\text{core}}_{i\leftarrow \mathrm{core}} &= \mathbf{1}\{i\in V_{\mathrm{core}}\}\sum_{j\in V_{\mathrm{core}}} D_{ji}. \label{eq:core_in}
    \end{align}
    \item Route conditioned demand for all nodes:
    \begin{align}
      a^{\text{all}}_{i\to \mathrm{cur}} &= \sum_{j\in V_{\mathrm{cur}}} D_{ij},&
      a^{\text{all}}_{i\leftarrow \mathrm{cur}} &= \sum_{j\in V_{\mathrm{cur}}} D_{ji}, \label{eq:all_cur}\\
      a^{\text{all}}_{i\to \mathrm{cmp}} &= \sum_{j\in V_{\mathrm{cmp}}} D_{ij},&
      a^{\text{all}}_{i\leftarrow \mathrm{cmp}} &= \sum_{j\in V_{\mathrm{cmp}}} D_{ji}. \label{eq:all_cmp}
    \end{align}
    \item Flags: one indicator each for current route membership $\mathbf{1}\{i\in V_{\mathrm{cur}}\}$, the fraction of completed routes that contain $i$ in $[0,1]$, and valid next node $\mathbf{1}\{i\in\mathcal{C}_t\}$.
\end{itemize}


\noindent This representation exposes the policy to demand patterns at multiple scales, from immediate next candidate nodes to the global network, enabling it to reason about ridership potential, transfers, and coverage, without requiring direct access to the OD matrix. Equations \eqref{eq:od_marginals} summarize the baseline incoming and outgoing demands at node $i$, independent of the current design. Equations \eqref{eq:cand_out} and \eqref{eq:cand_in} quantify the immediate marginal gain of selecting a candidate by measuring flows between a candidate and the current route. Equations \eqref{eq:core_out} and \eqref{eq:core_in} summarize how well a node is integrated with the already designed network. The route conditioned terms \eqref{eq:all_cur} and \eqref{eq:all_cmp} provide additional context i.e., a global view of demand relative to the current and completed routes.

\paragraph{Edge Connectivity ($\mathcal{I}$)} is the directed edge list that encodes graph topology for message passing. Since streets are bidirectional, we include both $(u,v)$ and $(v,u)$ i.e., $\mathcal{I}\;=\;\{(u,v)\in V\times V \mid \{u,v\}\in E\}$. 


\paragraph{Edge features ($Z\in\mathbb{R}^{|\mathcal{I}|\times 2}$)}
For each directed edge $(u,v)\in\mathcal{I}$, $Z$ stores a row with length and free flow speed.

\vspace{8pt}
\noindent All continuous features are min-max scaled to $[0,1]$ using network level bounds. Demand aggregates are computed from $D$ and scaled by the global reference $\max\!\bigl(\max_i\sum_j D_{ij},\ \max_j\sum_i D_{ji}\bigr)$. Terms that depend on $\mathcal{C}_t$ or $V_{\mathrm{core}}$ are set to zero outside those sets. Binary indicators take values in $\{0,1\}$.

\subsubsection{Action}

At each step, the agent selects an action $a_t \in \mathcal{A} = \{1,\ldots,n\}$ to append a node to the current route. The action space is naturally constrained by the network topology, i.e., at step $t$ only the admissible candidates $\mathcal{C}_t$ can be appended. Since $|\mathcal{C}_t|\ll n$ and this gap widens with network size, hence, we use action masking to concentrate probability mass on valid choices, which scales effectively when the space of invalid actions is large~\cite{huang2020closer}. We form a binary mask $m_t\in\{0,1\}^n$ over $\mathcal{C}_t$ with
\begin{equation}
    m_t[i] \;=\; \mathbf{1}\{\,i\in\mathcal{C}_t\,\}, \qquad i\in\{1,\ldots,n\}. 
\end{equation}
The policy then emits logits $z\in\mathbb{R}^n$ over $n$ and actions are sampled from a masked categorical distribution

\begin{equation}
\pi_{\theta}(a=i\mid s_t)\;=\;
\begin{cases}
\dfrac{\exp(z_i)}{\sum_{j:\ m_t[j]=1}\exp(z_j)} & \text{if } m_t[i]=1,\\[6pt]
0 & \text{if } m_t[i]=0.
\end{cases}
\label{eq:masked}
\end{equation}


This places zero probability on nodes outside $\mathcal{C}_t$. Node feasibility is also exposed in the state via valid-next flag in $X_t$, informing the policy and allowing it to condition on action constraints during training. When $\mathcal{C}_t=\varnothing$ (no valid extensions), current route construction terminates and the construction of the next route begins.

\subsubsection{Reward}
\label{subsec:reward}
TRNDP is a long horizon sequential decision problem that can require hundreds of decisions per episode even for modestly sized cities. A naive formulation that only scores the final network yields extremely sparse feedback and poor credit assignment~\cite{sutton2018reinforcement}. To mitigate this, we design a two-level reward function $\mathcal{R}(s_t, a_t)$ which combines a lightweight proxy available during route growth ($|r_k| < L_{\max}$) with a terminal reward from full traffic simulation at each route completion ($|r_k| = L_{\max}$ or terminal action). 


During the construction of a route $r_k$, the agent receives feedback based on the evolving network topology. Let $\Pi_{\text{built}} = \{r_1, \ldots, r_{k-1}\} \cup \{r_k\}$ denote all completed routes plus the current partial route. To encourage high-demand coverage, we compute the \emph{demand coverage potential} as:
\begin{equation}
    \Psi = \frac{\sum_{(i,j) \in \mathcal{P}_{\text{reachable}}} D_{ij}}{\sum_{i,j \in V} D_{ij}},
\end{equation}
where $\mathcal{P}_{\text{reachable}} = \{(i,j) : \exists \text{ a path from } i \text{ to } j \text{ in } \Pi_{\text{built}}\}$ denotes origin-destination pairs connected by the transit network. This measures the fraction of total demand that lies on designed routes.



Further, to discourage excessive duplication of service, we compute a \emph{route overlap} metric $\omega$. For each edge covered by at least one route, we measure its overlap depth, which ranges from $0$ (exclusive to one route) to $1$ (shared by all routes). The route overlap $\omega$ is the average depth across all covered edges. Combining these components, the incremental reward is:
\begin{equation}
    \mathcal{R}_{\text{partial}}(s_t, a_t) = \beta_0 \Psi - \beta_1 \omega,
\label{eq:partial}
\end{equation}
with $\beta_0 = 40$ and $\beta_1 = 20$. If a route terminates before reaching $L_{\max}$, it incurs an additional penalty $-\beta_2(1 - |r_k|/L_{\max})$ with $\beta_2 = 15$, proportional to how far it falls short of maximum length.


When route $r_k$ reaches $L_{\max}$ or terminates, we run a full traffic simulation over $T_{\text{sim}}$ steps with bus service frequencies determined by the maximum load principle (see supplementary material section~\ref{app:frequency}). The simulation tracks passenger journeys and provides feedback on design performance.

We compute the \emph{service rate}, which measures the fraction of servable demand that successfully boards buses:
\begin{equation}
    \sigma = \frac{N_{\text{boarded}}}{N_{\text{want}}},
    \label{eq:service_rate}
\end{equation}

 
where $N_{\text{boarded}}$ is the number of passengers who boarded buses during the simulation and $N_{\text{want}}$ denotes the potential riders. To penalize long passenger travel times, we compute the normalized average travel time:
\begin{equation}
    \tau = \min\!\left(\frac{1}{N_{\text{boarded}}} \sum_{p} \tau_p \;/\; 3600, \; 1\right),
\end{equation}
where $\tau_p$ denotes total travel time for passenger $p$ (includes both waiting and in-vehicle times). The terminal reward combines coverage potential, service effectiveness, travel efficiency, and overlap:
\begin{equation}
    \mathcal{R}_{\text{final}}(s_t, a_t) = \beta_3 \Psi + \beta_4 \sigma - \beta_5 \tau - \beta_6 \omega,
    \label{eq:final}
\end{equation}
with $\beta_3 = 30$, $\beta_4 = 15$, $\beta_5 = 15$, and $\beta_6 = 10$.

\vspace{8pt}
\noindent To stabilize their magnitudes during training, both partial and final rewards are normalized online~\cite{huang202237, rl_bag_of_tricks}. The weighting coefficients $(\beta_0,\ldots,\beta_6)$ are determined by hyper-parameter tuning. For more details on the policy networks and hyper-parameters, please refer to supplementary material section~\ref{app:policy} and section~\ref{app:hyperparams}, respectively.

\section{Experiment}
\label{sec:experiment}

\subsection{Setup}

We run all simulations in UXsim~\cite{seo2025uxsim}, a mesoscopic traffic simulator grounded in kinematic wave theory that uses Newell's simplified car-following model~\cite{newell2002simplified}. The simulator aggregates vehicles into platoons of size $\Delta n=5$ and advances dynamics with a discrete time step equal to the reaction-time parameter $\Delta t=1$ second, simulating to a horizon $T_{\text{sim}}=10{,}000$ steps ($\approx 2.7$ hours). To enable transit route network design, we extend UXsim with a reinforcement learning environment that adds a transit layer with mechanisms for buses and passengers boarding and alighting. The environment dispatches buses on routes designed by the agent with a capacity of $40$ passengers and a $60$\,s dwell time per stop, and allocates origin–destination demand between bus and car modes via a modal-split parameter $\alpha$. We train our RL policies on the Bloomington network for $2{,}000$ episodes, where each episode constructs a complete $16$-route transit network. Each route grows node by node until it reaches the maximum length of $L_{\max}=14$ nodes or until no valid one-hop extension exists.

We consider two modal-split scenarios: $\alpha=1.0$, where all served demand uses buses, and $\alpha=0.3$, a typical urban public transport share that assigns $30\%$ of demand to buses and the remainder to cars~\cite{buehler2019verkehrsverbund}. We also consider two route initialization schemes that differ only in how the first stop is chosen: \emph{transit center start}, where every route begins at a designated hub, and \emph{random start}, where the first stop is drawn uniformly from $V$. This yields four training configurations, resulting in four distinct RL policies.


\subsection{Baselines}
\label{subsec:baselines}
We compare our approach against five baseline methods that construct routes using different heuristic strategies.

\begin{itemize}
    \item \textbf{Real-world} baseline uses the $16$ existing bus routes in the Bloomington Transit network~\cite{bloomingtontransit_gtfs}, which were designed manually. This process typically involves various factors, such as historical data and operational constraints. This serves as a human-designed reference point that represents the current practice in transit network design.

    \item \textbf{Random walk} constructs routes by randomly selecting the next node at each step from the admissible candidate set $(\mathcal{C}_t)$. This baseline provides a stochastic reference to establish the benefits of more informed design strategies.

    \item \textbf{Greedy demand coverage} selects the candidate that maximizes immediate demand coverage at each step. For each candidate $i\in\mathcal{C}_t$, we compute the incremental demand score:
    \begin{equation}
        s(i) = \sum_{j \in r_k} D_{ij} + \sum_{j \in r_k} D_{ji},
    \end{equation}
    where $r_k$ denotes nodes already in the current route, and the candidate with the highest score is selected. This represents a natural demand-driven heuristic that prioritizes connections with higher traffic.

    \item \textbf{Greedy shortest path} constructs routes by selecting the nearest candidate at each step as measured by edge distance. This evaluates whether minimizing route length, a common objective in traditional routing problems, produces effective transit routes.

    
\end{itemize}

\subsection{Performance Metrics}
\label{subsec:metrics}

We evaluate the designs using seven simulation-driven metrics: service rate, wait time, transfer rate, and travel time assess passenger experience and service quality, while route efficiency, fleet size, and bus utilization capture operator-side efficiency. These metrics follow established frameworks for evaluating transit system performance~\cite{vuchic2007urban, vuchic2017urban}. At simulation end ($T_{\text{sim}}$ ), $N_{\text{comp}}$ denotes those who completed trips, $N_{\text{ongoing}}$ those currently traveling, and $N_{\text{waiting}}$ those still waiting at stops. Then $N_{\text{total}} = N_{\text{comp}} + N_{\text{ongoing}} + N_{\text{waiting}}$ denotes all passengers in the transit system.

\begin{figure*}[t!]
    \centering
    \includegraphics[width=0.70\linewidth]{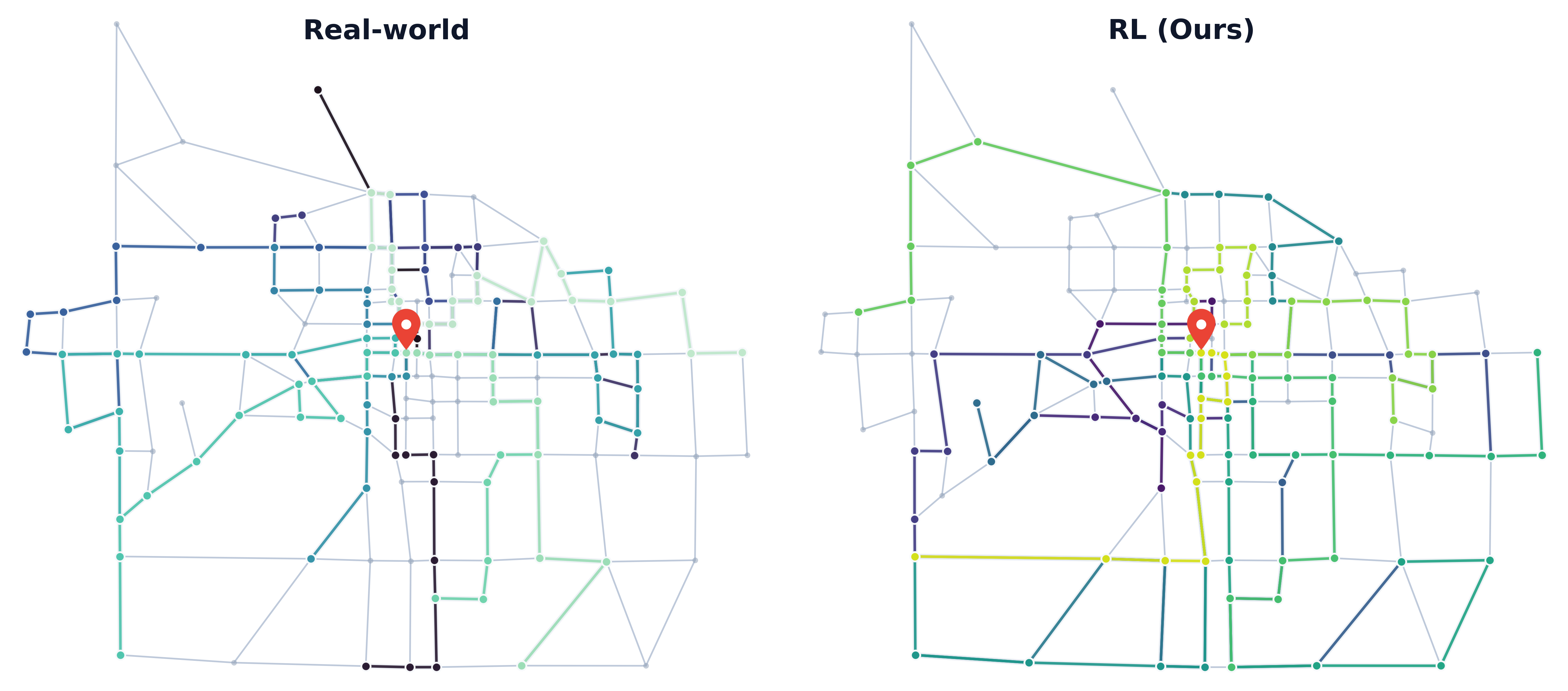}
    \vspace{-6pt}
    \caption{Comparison of real-world and RL-designed transit route networks in the Bloomington network at modal-split parameter $\alpha=0.3$. Gray nodes and edges show the base network; colored elements indicate transit routes. The RL design utilizes $143$ unique edges with $39$ shared by multiple routes, spreading service more broadly with less redundancy than the real-world network which has $135$ edges, among which $47$ are shared. The marker indicates the transit center (node $96$).}
    \label{fig:qualitative_route_analysis}
    \vspace{-10pt}
\end{figure*}

\begin{table}[t]
\centering
\renewcommand{\arraystretch}{1.1}
\scalebox{0.95}{
\begin{tabular}{lcc}
\multicolumn{3}{c}{$\bm{\alpha} = \mathbf{0.3}$} \\
\midrule
Metric & Real-world & RL (Ours) \\
\midrule
Service Rate (\%) $\uparrow$ & $42.28$ & $45.19$ \\
Wait Time (min) $\downarrow$ & $14.19$ & $10.32$ \\
Transfer Rate (\%) $\downarrow$ & $86.07$ & $86.90$ \\
Travel Time (min) $\downarrow$ & $48.12$ & $48.82$ \\
Route Efficiency (pax/km) $\uparrow$ & $13.17$ & $11.46$ \\
Fleet Size $\downarrow$ & $89$ & $92$ \\
Bus Utilization (\%) $\uparrow$ & $17.91$ & $18.78$ \\
\midrule
\addlinespace[0.5em]
\multicolumn{3}{c}{$\bm{\alpha} = \mathbf{1.0}$} \\
\midrule
Metric & Real-world & RL (Ours) \\
\midrule
Service Rate (\%) $\uparrow$ & $58.20$ & $73.10$ \\
Wait Time (min) $\downarrow$ & $15.87$ & $10.96$ \\
Transfer Rate (\%) $\downarrow$ & $82.14$ & $88.26$ \\
Travel Time (min) $\downarrow$ & $52.85$ & $53.04$ \\
Route Efficiency (pax/km) $\uparrow$ & $62.74$ & $72.06$ \\
Fleet Size $\downarrow$ & $281$ & $281$ \\
Bus Utilization (\%) $\uparrow$ & $32.15$ & $38.89$ \\
\bottomrule
\end{tabular}
}
\caption{Performance comparison between real-world and transit route networks designed by RL agent under two modal-split scenarios. Arrows indicate direction of improvement: $\uparrow$ means higher is better, $\downarrow$ means lower is better. At mixed-mode share ($\alpha=0.3$), RL improves service rate by $6.9\%$ and reduces wait times by $27.3\%$ while maintaining similar travel times, with $13\%$ lower route efficiency. At transit-only ($\alpha=1.0$), RL achieves substantial gains: $25.6\%$ higher service rate ($73.1\%$ vs $58.2\%$), $31\%$ shorter wait times, and $20.9\%$ better bus utilization. Despite inducing higher transfer rates, total travel time increases only marginally ($0.4$--$1.5\%$), indicating effective transfer coordination. Results for RL are averaged over $5$ independent designs whereas the real-world baseline represents a single fixed design.}
\label{tab:results_transit_center}
\vspace{-8pt}
\end{table}

\begin{itemize}
    \item \textbf{Service Rate} (\%) measures the fraction of potential demand that successfully boards, as defined in equation~\ref{eq:service_rate}. This directly quantifies the system's ability to convert reachable demand into actual ridership.

    \item \textbf{Wait Time} (minutes) is the average waiting time across all riders, computed as $\frac{1}{N_{\text{total}}} \sum_{p} \tau_p^{\text{wait}}$ where $\tau_p^{\text{wait}}$ is the waiting time for passenger $p$.

    \item \textbf{Transfer Rate} quantifies network connectivity by measuring the fraction of completed trips requiring transfers between routes:
            \begin{equation}
            \mathrm{Transfer~Rate~(\%)} = \frac{N_{\text{transfer}}}{N_{\text{comp}}} \times 100,
            \end{equation}
            where $N_{\text{transfer}}$ is the count of trips with at least one transfer. 

    \item \textbf{Travel Time} captures total trip duration, including both wait and in-vehicle time, computed for passengers who complete trips or are mid-journey at simulation end:
            \begin{equation}
            \mathrm{Travel~Time~(min)} = \frac{1}{N_{\text{served}}} \sum_{p} \bigl(\tau_p^{\text{wait}} + \tau_p^{\text{move}}\bigr),
            \end{equation}
            where $\tau_p^{\text{move}}$ is in-vehicle travel time for passenger $p$.

    \item \textbf{Route Efficiency} (pax/km) measures passengers served per unit of infrastructure as the ratio of completed trips $N_{\text{comp}}$ to total route length $L_{\text{total}}$ in kilometers.

    \item \textbf{Fleet Size} is the total bus count deployed across all routes.
    
    \item \textbf{Bus Utilization} ($\%$) measures average capacity usage across the fleet.
\end{itemize}

\subsection{Results}
\label{subsec:results}

\begin{figure*}[t!]
    \centering
    \includegraphics[width=0.93\linewidth]
    {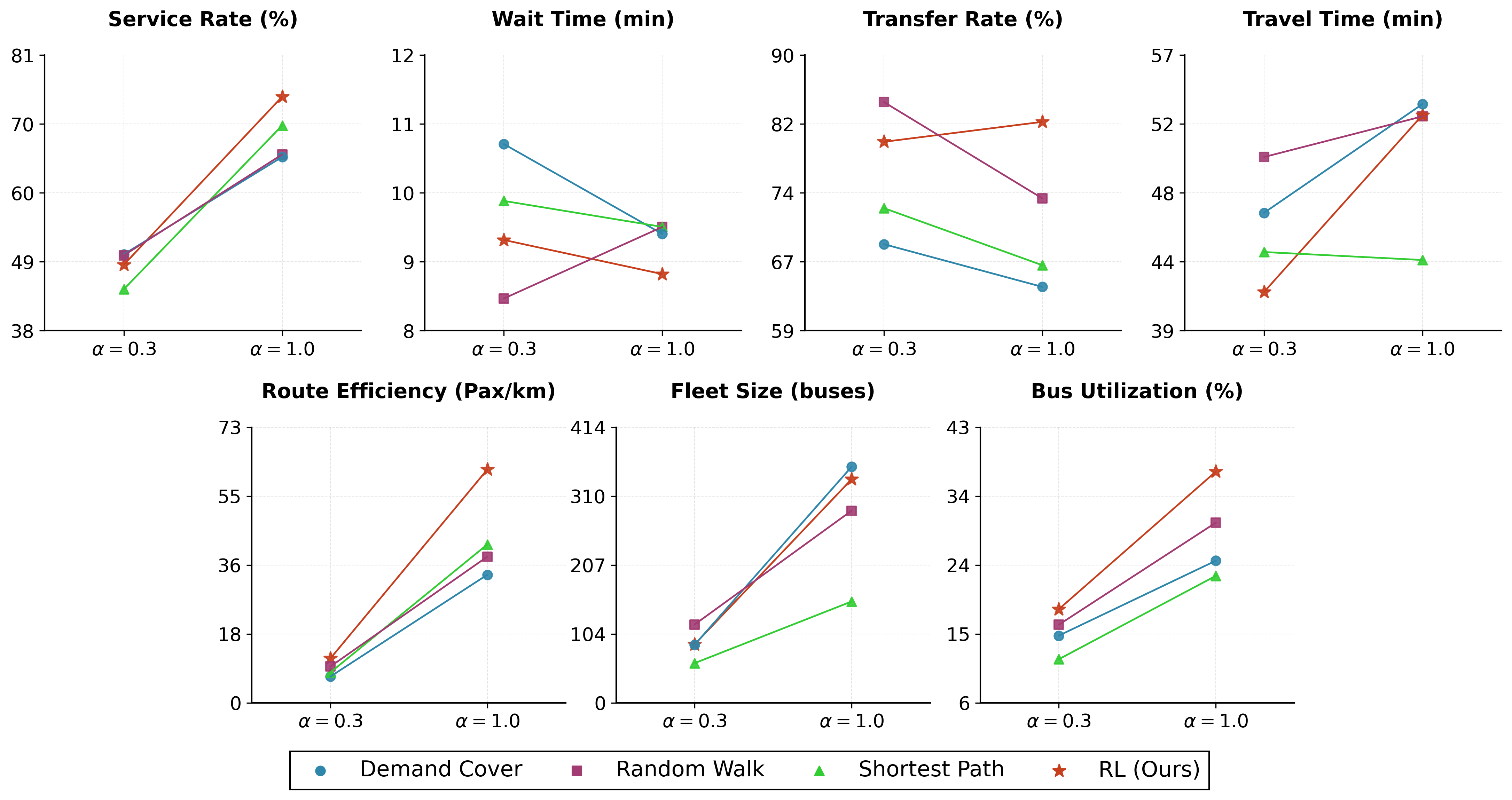}
    \vspace{-5pt}
    \caption{Comparison of transit network designs with random initialization at modal-split $\alpha=0.3$ and $\alpha=1.0$. The top four metrics reflect passenger experience, while the bottom three reflect operator performance. RL-designed networks achieve $68.8\%$ higher route efficiency than greedy demand coverage at $\alpha=0.3$ and deliver the shortest travel times ($41.53$ minutes). At $\alpha=1.0$, RL achieves the highest service rate ($74.5\%$) and $47.3\%$ higher route efficiency than shortest path. No single baseline dominates across all metrics, reflecting inherent trade-offs in multi-objective transit design. Results are averaged over $5$ independent designs.}
    \vspace{-10pt}
    \label{fig:random_init}
\end{figure*}

We evaluate our trained policies against the baselines across two initializations (transit center and random) and two modal-splits ($\alpha = 0.3$ and $\alpha = 1.0$). All methods respect the constraints defined in Sec.~\ref{sec:method} and are simulated with identical parameters and frequency setting rule.

Table~\ref{tab:results_transit_center} compares networks designed by the RL policy with the real-world routes, using the transit center initialization scheme to match the real-world configuration. At $\alpha = 0.3$, the RL-designed network achieves a $6.9\%$ increase in service rate (from $42.28\%$ to $45.19\%$) and reduces average wait time by $27.3\%$ (from $14.19$ to $10.32$ minutes), while maintaining nearly identical travel times ($48.12$ vs $48.82$ minutes). However, this design exhibits $13\%$ lower route efficiency ($11.46$ vs $13.17$ passengers/ km) and requires three additional buses ($92$ vs $89$). At $\alpha = 1.0$, performance gains become substantially more pronounced: the RL-designed network achieves a $25.6\%$ increase in service rate (from $58.20\%$ to $73.10\%$), reduces wait time by $30.9\%$ (from $15.87$ to $10.96$ minutes), and delivers $14.9\%$ higher route efficiency ($72.06$ vs $62.74$ pax/km) and $21.0\%$ better bus utilization ($38.89\%$ vs $32.15\%$) with the same fleet size of $281$ buses. These results show that performance gains scale substantially as transit mode share increases. The trade-offs include slightly higher transfer rates (increasing by $0.8$--$6.1$ percentage points) and, at $\alpha=0.3$, lower route efficiency.

Qualitative analysis of the routes (at $\alpha=1.0$ with transit center initialization) reveals distinct differences in network organization between the RL-designed and real-world networks, as shown in Figure~\ref{fig:qualitative_route_analysis}. Both designs achieve similar coverage, reaching $117$ unique nodes (RL) compared to $114$ (real-world). However, the networks serve partially different geographic areas: the RL network reaches $27$ nodes not served by the real-world system, while the real-world network exclusively serves $24$ different nodes. This suggests the RL agent prioritizes different demand patterns than those emphasized in manual planning. Examining network topology more closely, the RL design utilizes $143$ unique edges compared to $135$ in the real-world network, while exhibiting less route overlap: only $39$ edges are shared by multiple routes compared to $47$ in the real-world design. Additionally, the real-world network concentrates service more heavily, with $22.8\%$ of nodes served by three or more routes, compared to $19.7\%$ in the RL design. This indicates that the RL policy spreads service more broadly across the network with less redundancy, while the real-world design concentrates capacity along fewer shared segments.

Figure~\ref{fig:random_init} compares the RL-designed networks against three heuristic baselines under random initialization. At $\alpha=0.3$, greedy demand coverage achieves the highest service rate ($49.9\%$) but delivers route efficiency of only $6.96$ passengers/ km. RL policy achieves a competitive service rate of $48.3\%$ with $68.8\%$ higher route efficiency ($11.75$ vs $6.96$ pax/km) and the shortest travel time ($41.53$ minutes), which are $5.9\%$ lower than even the shortest path baseline ($44.14$ minutes). This demonstrates that distance minimization creates spatially compact but poorly coordinated networks that fail to account for network-wide effects such as congestion and transfer coordination. Random walk exhibits competitive service rates ($49.8\%$) but requires $33.9\%$ more buses ($117.6$ vs $87.8$) and produces the longest travel time ($50.34$ minutes). At $\alpha=1.0$, RL policy achieves the highest service rate ($74.5\%$), outperforming shortest path ($70.0\%$) by $6.4\%$, random walk ($65.5\%$) by $13.7\%$, and demand coverage ($65.1\%$) by $14.4\%$. RL delivers $47.3\%$ higher route efficiency than shortest path ($61.82$ vs $41.95$ pax/km), the best bus utilization ($37.04\%$), and lowest wait time ($8.82$ minutes). However, RL networks exhibit higher travel time ($53.08$ minutes) comparable to random walk ($53.0$ minutes) but $21.7\%$ higher than shortest path ($43.62$ minutes), reflecting increased transfers ($82.49\%$) as the policy prioritizes broader coverage and throughput over trip duration. No single baseline dominates across all metrics: demand coverage maximizes service rate but sacrifices efficiency; shortest path minimizes fleet size but compromises coverage; random walk spreads service broadly but lacks coordination. RL policy learns to balance these competing objectives, demonstrating competitive or superior performance on most metrics across both demand scenarios.

\section{Conclusion and Future Work}
\label{sec:conclusion}
In this work, we introduced an end-to-end RL framework for transit route network design that integrates a graph attention policy and a two-level reward structure with a mesoscopic traffic simulator. We released a real-world dataset from Bloomington, Indiana with a topologically accurate road network ($143$ nodes, $243$ edges), census-based origin–destination demand, and $16$ existing transit routes, providing a realistic testbed that captures the rich structure and demand patterns absent in synthetic benchmarks. On this network, the learned policies substantially improve passenger metrics and, particularly at high transit mode share ($\alpha = 1.0$), operator metrics relative to the current system. The RL design increases service rate from $58.2\%$ to $73.1\%$, reduces average wait time by $30.9\%$, and increases bus utilization by $21.0\%$ while maintaining comparable travel times. Under $\alpha = 0.3$ (mixed-mode share) with transit center initialization, our approach improves service rate by $6.9\%$ and reduces wait times by $27.3\%$ relative to the real-world design. In random initialization experiments, it achieves $68.8\%$ higher route efficiency than greedy demand coverage and $5.9\%$ lower travel times than shortest path construction, illustrating that the learned policy finds favorable trade-offs that no single heuristic baseline attains. These gains translate into more accessible and resource-efficient transit service, demonstrating the potential of data-driven methods to advance beyond traditional manual planning approaches and reduce reliance on hand-crafted heuristics in urban transit design. 


Despite promising results, there are several limitations. First, evaluation is restricted to a single mid-sized city and the framework assumes static peak-hour demand rather than time-varying patterns and stochastic fluctuations.
In the future, we would like to test our approach on more complex city networks that are imbued with real-world traffic data and advanced traffic simulation techniques~\cite{Guo2024Simulation,Li2017CityFlowRecon,Lin2022GCGRNN,Lin2019Compress,Li2018CityEstIET,Li2017CitySparseITSM,Wilkie2015Virtual,Chao2020Survey}. Furthermore, we would like to test the integration of work with future mobility efforts, especially the emergence of mixed traffic dynamics~\cite{Pan2025Review,Liu2025Large,Islam2025Heterogeneous,Fan2025OD,Wang2024Intersection,Wang2024Privacy,Poudel2024CARL,Poudel2024EnduRL,Villarreal2024Eco,Villarreal2023Pixel,Villarreal2023Chat}.


\bibliographystyle{IEEEtran}
\bibliography{main}

\clearpage
\setcounter{page}{1}

\section{Definitions}
\label{sec:definitions}

We formalize the components of the transit network and the design process below:
\vspace{11pt}

\noindent\textbf{Definition 1 (Road Network).} A road network is an undirected graph $G = (V, E)$ where $V = \{v_1, \ldots, v_n\}$ is the set of $n$ nodes representing intersections or stops, and $E \subseteq \{\{u,v\} \mid u,v \in V, u \neq v\}$ is the set of edges representing bidirectional road segments. Each edge $e = \{u,v\} \in E$ has an associated length $\ell_e > 0$ and free-flow speed $s_e > 0$. The neighborhood of node $v$ is $\mathcal{N}(v) = \{u \in V \mid \{u,v\} \in E\}$.

\vspace{11pt}
\noindent\textbf{Definition 2 (Transit Route).} A transit route $r_k$ is a simple path in $G$ defined as an ordered sequence of distinct nodes:
\[
r_k = (v_{k,1}, v_{k,2}, \ldots, v_{k,\ell_k})
\]
where $\ell_k \geq 2$ is the route length, $\{v_{k,i}, v_{k,i+1}\} \in E$ for all $i \in \{1, \ldots, \ell_k-1\}$, and $v_{k,i} \neq v_{k,j}$ for $i \neq j$. Routes are bidirectional, with vehicles traversing both $v_{k,1} \to v_{k,\ell_k}$ and $v_{k,\ell_k} \to v_{k,1}$.

\vspace{11pt}
\noindent\textbf{Definition 3 (Transit Network).} A transit network $\Pi = \{r_1, \ldots, r_K\}$ is a collection of $K$ routes operating on $G$. The network must satisfy constraints such as route length: $L_{\min} \leq |r_k| \leq L_{\max}$ for all $k \in \{1, \ldots, K\}$.

\vspace{11pt}
\noindent\textbf{Definition 4 (Stops and Frequency of Service).} For route $r_k$ with stop spacing parameter $s_k \geq 1$, the stop set is:
\[
\mathcal{S}(r_k) = \{v_{k,1+js_k} \mid j = 0, 1, \ldots, \lfloor(\ell_k - 1)/s_k\rfloor\}
\]
Each route operates with frequency $\mathcal{F}_k \in \mathbb{N}_{>0}$ vehicles per hour in each direction. The tuple $(r_k, s_k, \mathcal{F}_k)$ fully specifies the operational characteristics of route $k$.

\vspace{11pt}
\noindent\textbf{Definition 5 (Origin-Destination Demand).} The travel demand is represented by matrix $D \in \mathbb{R}_{\geq 0}^{n \times n}$ where $D_{ij}$ denotes passenger flow (trips/hour) from origin $i$ to destination $j$. The modal split parameter $\alpha \in [0,1]$ determines the fraction of demand using transit, yielding effective transit demand $D^{\text{transit}}_{ij} = \alpha D_{ij}$.

\vspace{11pt}
\noindent\textbf{Definition 6 (Sequential Route Construction).} Routes are constructed incrementally through a sequence of decisions. At step $t$, the partial route is $r^{(t)} = (v_1, \ldots, v_t)$ with frontier node $v_t$. The admissible extension set is:
\[
\mathcal{C}_t = \{u \in \mathcal{N}(v_t) \mid u \notin \{v_1, \ldots, v_t\}\}
\]
representing valid next nodes that maintain the simple path property.

\vspace{11pt}
\noindent\textbf{Definition 7 (Transfer).} A transfer occurs when a passenger journey from origin $i$ to destination $j$ requires traversing multiple routes. Let $\mathcal{P}_{ij}$ denote the set of feasible paths from $i$ to $j$ using the transit network. A path $p \in \mathcal{P}_{ij}$ requiring $m > 1$ distinct routes incurs $m-1$ transfers.

\section{Real-world Network and Data Processing}
\label{app:data_processing}

We introduce a novel city-scale transit network dataset for Bloomington, Indiana. Unlike existing benchmark networks from literature, our dataset uniquely captures real-world aspects in three dimensions: (i) the underlying transportation network, (ii) travel demand derived from census data, and (iii) transit routes currently operating in the city. 

\subsection{Network Structure}
The network consists of $143$ nodes and $243$ bidirectional edges, spanning a coverage area of approximately $152.3~\text{km}^{2}$. The network topology was derived from actual road infrastructure with several practical assumptions made to balance modeling fidelity with simulation efficiency.

\begin{itemize}
    \item \textbf{2D Representation}: 3D infrastructure elements such as tunnels, overpasses, and underpasses are modeled as planar (2D) connections.

    \item \textbf{Edge Geometry}: All edges are represented as bidirectional. When two parallel one-way streets exist next to each other, they are consolidated into single a bidirectional edge positioned at their centerline. Further, edges follow shortest-distance connections between nodes rather than exact street curvatures; the length differences are negligible for the scale of analysis.

    \item \textbf{Speed Assignment}: All edges are assigned a uniform free-flow speed of $16.67$ m/s ($60$ km/h or $37$ miles/hr), reflecting typical urban traffic speed limit.

    \item \textbf{Highway Exclusion}: Interstate $69$ highway segments within the city limits were excluded from the transit network, as city transit primarily serves local destinations and highways lack appropriate passenger access points. The bus routes currently operating in the city also do not utilize the highway.

    \item \textbf{Shared Routes}: Although existing bus routes may have slightly different paths for outbound and inbound directions, to reduce complexity, we model them as identical paths, manually selecting the most appropriate nodes (based on access, length, and community served) to maintain essential connectivity.
\end{itemize}

\subsection{Coordinate Transformation}
The raw geospatial data uses geographic coordinates (latitude and longitude) in angular units. However, operations in our processing pipeline such as calculating census block centroids and edge distances require a flat, Cartesian plane for accurate results. Therefore, the coordinates were transformed to a Cartesian coordinate system using the Universal Transverse Mercator (UTM) Zone 16N projection, which covers Indiana including Bloomington. The UTM projection preserves local angles and shapes while providing true metric distances and areas with minimal distortion, making it ideal for our case.

\subsection{Demand generation}
The primary demand component was obtained from commuting trips captured in the $2022$ LEHD Origin-Destination Employment Statistics (LODES) from the U.S. Census Bureau~\cite{uscensus_lehd_lodes}, which provides flows between census blocks with home locations as origins and work locations as destinations. We utilize block-level census data~\cite{uscensus_tiger_shapefiles} and processed a total of $2,399$ census blocks within Monroe County, Indiana (FIPS code $105$, GEOID $18105$), which was further reduced to $1,475$ blocks to confine the area of interest to the vicinity of Bloomington. For each census block, its centroid was calculated and the demand origins and destinations within that block were assigned to a node in the network nearest to the centroid.

Because the LODES data excludes trips for non-commuting purposes such as school and shopping (typically classified as home-based other or non-home-based in transportation research), to account for this mixed composition of traffic, we scale the commuting flows by $150\%$, consistent with typical values ranging from $100\%-200\%$ depending on time of day~\cite{mcguckin2018summary}. Further, to express demand on an hourly basis, we adopt a peak‑hour share of $11\%$ of daily traffic, which lies within the typical $6\%-12\%$ range~\cite{roess2011traffic}. The resulting origin–destination (OD) demand matrix contains $5,738$ pairs, with a maximum origin demand of $344$ vehicles per hour and a maximum destination demand of $1,681$ vehicles per hour.

\subsection{Existing bus routes}
The Bloomington Transit system~\cite{bloomingtontransit_gtfs, transitland_bloomingtontransit} operates $16$ bus routes which map to an average of $14.2$ nodes per route in our network (ranging from $8$ to $24$ nodes). 

\vspace{10pt}
\noindent We make the complete dataset publicly available (as three csv and one json files for nodes, edges, demand, and existing routes) to support future research in related areas. Our dataset and code can be found in the repository: \url{https://anonymous.4open.science/r/Transit_Design_CVPR-F0DB}


\section{Size of the search space}
\label{app:searchspace}
The scope of the design problem is to choose $R=16$ routes, each visiting $14$ distinct nodes, on the Bloomington network graph $G=(V, E)$. Computing the exact size of the search space is computationally intensive, so we estimate it by approximating the number of simple paths (no repeated nodes) consisting of $L=13$ edges. 

The graph has $|V|=143$ and $|E|=243$, which gives the average degree
\[
d \;=\; \frac{2|E|}{|V|} \;=\; \frac{486}{143} \;\approx\; 3.40.
\]
For simple paths in sparse graphs, we approximate the number of candidate routes by accounting for the constraint that nodes cannot repeat. Starting from any of $|V|$ nodes, the first step has approximately $d$ choices, and each subsequent step has approximately $d-1$ choices (excluding the node just visited):
\[
P \;\approx\; |V| \cdot d \cdot (d-1)^{L-1}.
\]
Treating a design as an ordered list of $R=16$ routes, the total search space is
\[
S \;=\; P^{R} \;\approx\; \bigl(|V| \cdot d \cdot (d-1)^{L-1}\bigr)^{16}.
\]

Numerically,
\[
(d-1)^{12} = (2.40)^{12} \approx 3.65\times 10^{4},~
P \approx 1.78\times 10^{7},
\]
\[
S \approx 9.5\times 10^{115}.
\]

\noindent The magnitude of $S$ makes an exhaustive or near-exhaustive search infeasible and motivates alternative approaches that exploit the problem's structure. We note that this approximation does not adjust for overlapping edges between multiple routes.

\begin{figure*}[t!]
    \centering
    \includegraphics[width=0.95\linewidth]{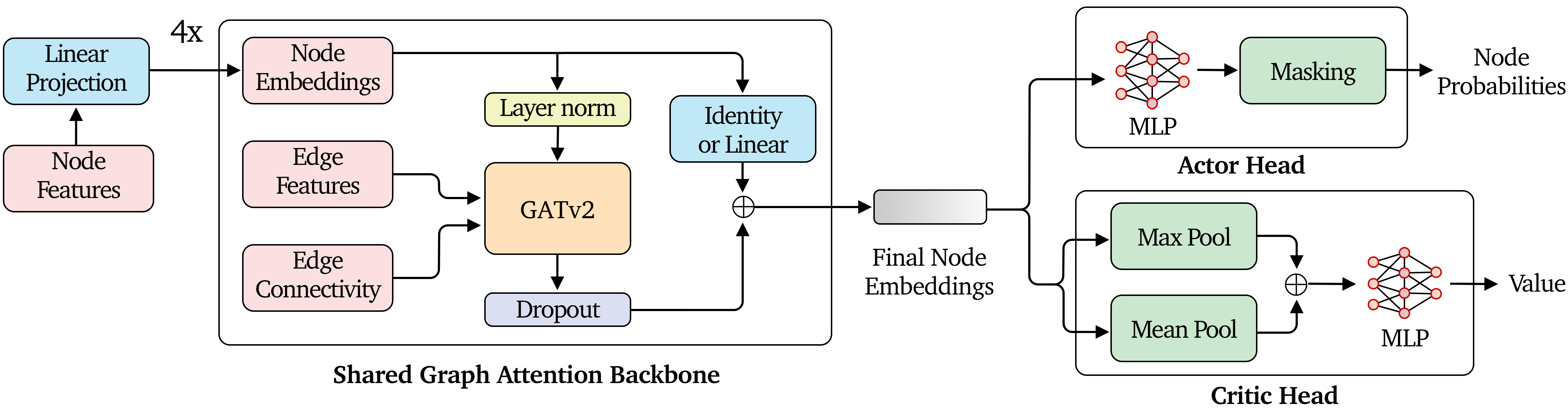}
    \vspace{-5pt}
    \caption{Overview of the policy network. A shared graph attention backbone maps $16$ node features, $2$ edge features, and edge connectivity to final node embeddings of size $64$. First, the node features are projected to $64$ channels. Then the representation passes through a stack of four GATv2 blocks in sequence with the output of attention heads averaged (instead of concatenated) to control channel growth. Each block uses pre-layer normalization, a GATv2 attention layer with dropout, and a residual path that copies the input when widths match or applies a linear projection when they differ. The \textbf{actor head} is an MLP with three hidden layers of widths $256$, $128$, and $64$ that produces a score per node, followed by a feasibility mask to obtain probabilities over valid nodes. The \textbf{critic head} concatenates global mean and global max pooling of the node embeddings, then feeds the result to an MLP with three hidden layers of widths $256$, $128$, and $64$ to predict a scalar value per graph.}
    \vspace{-6pt}
    \label{fig:policy_network}
\end{figure*}


\section{Frequency of service (FOS) assignment}
\label{app:frequency}

Frequency of service (FOS) setting is an integral component of the Transit Route Network Design Problem, as it directly affects both passenger satisfaction (via wait and travel times) and operator costs (e.g., fleet size). FOS measures the number of buses dispatched per hour on a route to maintain adequate service. Learning FOS on a sequential route design (either jointly during route construction or at episode end) yields inconsistent or mis-guided learning signals. For example, during initial stages of the route construction process, the route might just be two nodes \([1 \to 5]\), this short segment only serves a single O-D pair. The agent may learn that setting a high frequency is beneficial for reducing wait times, etc. But later, when routes are longer, high frequencies might induce over-servicing and increased costs; i.e., a frequency of 6 buses/hour is a bad choice for a 2-node route but might be an excellent choice for a 10-node route. This inconsistency makes it challenging for the RL agent to learn a coherent policy, as it would have to adapt to completely different frequency preferences at each step of route construction, which is extremely difficult. We instead compute FOS deterministically post-route design using a max-load principle~\cite{ceder2016public, furth1981setting}. For each route $k$, frequency is set to ensure the peak passenger load on the busiest segment does not exceed comfortable capacity: 

\begin{equation}
\mathcal{F}_k = \left\lceil \frac{\mathcal{Q}{k, \max}^\text{norm}}{\delta_{\max} \cdot \mathcal{C}_k} \right\rceil
\label{eq:fos}
\end{equation}

where $\mathcal{F}_k$ is the frequency for route $k$, $\mathcal{Q}_{k, \max}^\text{norm}$ is the maximum normalized segment load, $\delta_{\max}$ is the comfort threshold (maximum desired load factor, e.g., $0.8$ for $80\%$ capacity), and $\mathcal{C}_k$ is bus capacity. To compute $\mathcal{Q}_{k, \max}^\text{norm}$, we account for both direct trips (origin and destination on route $k$) and transfer trips (using route $k$ as part of a multi-route journey). During normalization, we divide each segment's load by the number of overlapping routes to prevent overestimation of frequency requirements when multiple routes serve the same segment. The maximum normalized load across all segments of route $k$ determines $\mathcal{F}_k$. While the RL agent has no direct control over frequency assignment, it indirectly influences it through its route design decisions: for example, routes serving high-demand corridors receive higher frequencies, whereas when multiple routes overlap a segment, the normalized load decreases and each route receives proportionally lower frequency.


\section{Policy Network}
\label{app:policy} 

The policy network consists of a shared graph attention backbone and two separate heads for the actor and critic, as shown in Figure~\ref{fig:policy_network}. To make the policy scalable across cities of different sizes, we design the architecture so that the number of trainable parameters is independent of the node count $n$, while the forward computational cost grows linearly with it. This is achieved by: (i) using a shared Graph Attention Networks v2 (GATv2)~\cite{brody2021attentive} encoder that produces fixed-width node embeddings for any graph size, (ii) implementing an actor that assigns a score to each node with one shared MLP. For the critic, we apply mean and max pooling to the final node embeddings and concatenate it before passing as input. Together, these design choices keep the actor permutation equivariant, the critic permutation invariant i.e., ensuring the policy remains stable under node reindexing, and the parameter count independent of the number of nodes $n$ in the input graph.

\paragraph{Shared GAT backbone}
At time $t$, the input is $(X_t, \mathcal{I}, Z)$ where $X_t\in\mathbb{R}^{n\times 16}$ are node features, $\mathcal{I}$ represents edge connectivity as a directed edge list, and $Z\in\mathbb{R}^{|\mathcal{I}|\times 2}$ are edge attributes. A linear projection maps node features to $64$ channels, then a stack of four \textsc{GATv2} blocks produces final node embeddings
\[
H = \textsc{GATv2Stack}\bigl(\textsc{Proj}(X_t), \mathcal{I}, Z\bigr) \in \mathbb{R}^{n\times 64}.
\]
Compared with the original GAT~\cite{velivckovic2017graph}, \textsc{GATv2} employs a dynamic attention mechanism that depends jointly and nonlinearly on both endpoints and is strictly more expressive~\cite{brody2021attentive}. It also conditions on edge attributes, so the edge features (length and free-flow speed) inform message passing and improve modeling of link-dependent interactions. Each block applies pre-layer normalization, a \textsc{GATv2} attention layer with averaged heads, attention and feature dropout, and an identity skip connection when input and output widths match, or a linear projection when they differ.Block $1$ outputs width $128$ with $8$ heads and uses a projection on the skip ($64\!\to\!128$); Block $2$ outputs width $128$ with $8$ heads and keeps identity ($128\!\to\!128$); Block $3$ outputs width $64$ with $4$ heads and uses a projection ($128\!\to\!64$); Block $4$ outputs width $64$ with $4$ heads and keeps identity ($64\!\to\!64$). 

\paragraph{Actor}
The actor head uses a pointer-style scoring mechanism~\cite{vinyals2015pointer,yang2022graph}. A shared multi-layer perceptron (MLP) scores all nodes in parallel, applies a feasibility mask to restrict the distribution to admissible candidates, and selects exactly one node per step. Unlike decoder-based pointer networks that use a recurrence to emit an ordered sequence of pointers, our single-step selection naturally aligns with RL's sequential route construction where decisions are made one node at a time. The actor MLP consists of three hidden layers of widths $256$, $128$, and $64$ that produces a scalar score per node, with weights shared across all nodes. A feasibility mask from the environment restricts probability mass only to the admissible candidates that can extend the current route. Node features $X_t$ include frontier membership and valid next indicators, so the scorer implicitly conditions on the evolving route context.

\paragraph{Critic}
The critic head concatenates global mean and global max pooling of the node embeddings, then uses an MLP with three hidden layers of widths $256$, $128$, and $64$ to predict a scalar value $V(G)$ per graph.

\vspace{8pt}
\noindent The backbone is computed once per graph to obtain $H$, after which the actor produces per-node scores in a single pass and the critic aggregates with constant-time pooling. This yields a forward pass that scales as $\mathcal{O}(B|\mathcal{I}|d + nd)$ for $B$ blocks of width $d$. Implementation-wise, all linear layers use orthogonal initialization, and dropout is applied on attention coefficients and block outputs in the backbone, as well as in the hidden layers of both heads for regularization.

\section{Training and Hyperparameters}
\label{app:hyperparams}

Training occurs over a total of \(2000\) episodes (around \(0.4\)M max simulation timesteps), on an Intel Core i\(9\)-\(14900\)K processor and an NVIDIA RTX \(4090\) GPU. The total wall-clock training time is approximately \(48\) hours. Table~\ref{table:params} summarizes the core hyperparameters covering the simulation settings, policy architecture, and PPO training.

\begin{table}[h!]
    \centering
    \setlength{\tabcolsep}{5.5pt}
    \renewcommand{\arraystretch}{1.1}
    \begin{tabular}{llr}
        \toprule
        \textbf{Category} & \textbf{Parameter} & \textbf{Value} \\
        \toprule
        \multirow{9}{*}{Simulation}
        & Simulation horizon (\(T_{\text{sim}}\)) & \(10, 000\) steps \\
        & Time step (\(\Delta t\)) & \(1\) s \\
        & Stop spacing & \(1\) node\\
        & Bus capacity & \(40\) passengers\\
        & Stop duration & \(60\) s \\
        & Modal split (\(\alpha\)) & \(0.3\) \\
        & Number of routes & \(16\) \\
        & Max route length & \(14\) \\
        & Comfort threshold & \(1.0\) \\
        \midrule
        \multirow{9}{*}{Policy}
        & \# Node features & \(16\) \\
        & \# Edge features & \(2\) \\
        & \# GATv$2$ blocks & \(4\) \\
        & \# Channels & \([128,128,64,64]\) \\
        & \# Attention heads & \([8,8,4,4]\) \\
        & Activation & \texttt{tanh} \\
        & Actor head layers & \([256,128,64]\) \\
        & Critic head layers & \([256,128,64]\) \\
        & Attention dropout & \([0, 0.05, 0.10, 0.10]\) \\
        \midrule
        \multirow{10}{*}{PPO}
        & Max episode steps & \(224\) \\
        & Learning rate & \(5\times10^{-4}\) \\
        & Discount (\(\gamma\)) & \(0.99\) \\
        & GAE (\(\lambda\)) & \(0.95\) \\
        & Epochs per update (\(K\)) & \(8\) \\
        & Mini batch size & \(16\) \\
        & Update frequency & \(128\) \\
        & Policy clip (\(\epsilon\)) & \(0.2\) \\
        & Entropy coef & \(0.02\) \\
        & Max grad norm & \(0.5\) \\
        \bottomrule
    \end{tabular}
    \caption{Core environment settings and hyper-parameters used.}
    \label{table:params}
\end{table}

\end{document}